\documentclass[manuscript,screen,nonacm]{acmart}
\setcopyright{none}
\AtBeginDocument{%
  }

\usepackage{subfig}
\usepackage{threeparttable}
\usepackage{multirow}
\usepackage{booktabs}
\usepackage[most]{tcolorbox}
\usepackage{array, xcolor}
\usepackage{pifont}
\usepackage{colortbl}
\usepackage{adjustbox}
\usepackage{fontawesome}
\usepackage[colorinlistoftodos]{todonotes}
\begin{document}

%%
%% The "title" command has an optional parameter,
%% allowing the author to define a "short title" to be used in page headers.
\title{Pub-Guard-LLM: Reliable and Fair Retraction Risk Assessment for Biomedical Publications}

%%
%% The "author" command and its associated commands are used to define
%% the authors and their affiliations.
%% Of note is the shared affiliation of the first two authors, and the
%% "authornote" and "authornotemark" commands
%% used to denote shared contribution to the research.
% Preprint authors: order and affiliations from supplied screenshot.
% Contact email for corresponding author.
\author{Lihu Chen}
\email{lihu.chen@imperial.ac.uk}
\affiliation{%
  \institution{Imperial College London}
  \city{London}
  \country{United Kingdom}
}
\author{Hamed Ayoobi}
\affiliation{%
  \institution{University of Groningen}
  \city{Groningen}
  \country{Netherlands}
}
\author{Shuojie Fu}
\affiliation{%
  \institution{Imperial College London}
  \city{London}
  \country{United Kingdom}
}
\author{Gabriel Freedman}
\affiliation{%
  \institution{Imperial College London}
  \city{London}
  \country{United Kingdom}
}
\author{Cemre Zor}
\affiliation{%
  \institution{Amazon}
  \country{United Kingdom}
}
\author{Guy Martin}
\affiliation{%
  \institution{National Health Service}
  \country{United Kingdom}
}
\author{James Kinross}
\affiliation{%
  \institution{Imperial College London}
  \city{London}
  \country{United Kingdom}
}
\author{Uddhav Vaghela}
\affiliation{%
  \institution{Imperial College London}
  \city{London}
  \country{United Kingdom}
}
\author{Ovidiu Serban}
\affiliation{%
  \institution{Imperial College London}
  \city{London}
  \country{United Kingdom}
}
\author{Francesca Toni}
\affiliation{%
  \institution{Imperial College London}
  \city{London}
  \country{United Kingdom}
}
\renewcommand{\shortauthors}{Chen et al.}

%%
%% The abstract is a short summary of the work to be presented in the
%% article.
\begin{abstract}
A growing number of scientific articles are retracted each year. Many of these articles continue to be cited and influence research and clinical decisions, posing serious risks.
In this paper, we propose \texttt{Pub-Guard-LLM}, a large language model-based system for assessing retraction risks in biomedical articles. The system predicts retraction risks and provides textual explanations. We introduce an open-source benchmark, \emph{PubMed Retraction}, containing over 12K real-world biomedical articles with metadata and retraction labels. Our results show that \texttt{Pub-Guard-LLM} outperforms various baselines.
We further show that adding metadata, such as author, institution, and journal, improves prediction performance.
However, metadata can also introduce bias. We study metadata-related bias in retraction risk predictions using a framework for transparent bias detection. 
To reduce bias, we explore several methods, including retrieval-augmented generation, counterfactual fine-tuning, debate, and argumentation-based approaches. Our results reveal a trade-off between prediction performance and metadata bias.
\texttt{Pub-Guard-LLM} provides a step towards reliable and fair retraction risk assessment for biomedical research.
\end{abstract}


%%
%% Keywords. The author(s) should pick words that accurately describe
%% the work being presented. Separate the keywords with commas.
%% A "teaser" image appears between the author and affiliation
%% information and the body of the document, and typically spans the
%% page.
% \begin{teaserfigure}
%  \includegraphics[width=\textwidth]{sampleteaser}
%   \caption{Seattle Mariners at Spring Training, 2010.}
%   \Description{Enjoying the baseball game from the third-base
%   seats. Ichiro Suzuki preparing to bat.}
%   \label{fig:teaser}
% \end{teaserfigure}

% Preprint: no journal received / accepted dates

%%
%% This command processes the author and affiliation and title
%% information and builds the first part of the formatted document.
\maketitle

\section{Introduction}
A considerable proportion of published scientific articles are retracted. It is estimated that between 1.5 and 2\% of all scientific papers published in 2022 closely resemble content from paper-mill products and retracted articles \citep{fakepaper2023nature}. In 2023, about 8,000 papers were retracted from Hindawi journals due to their fraudulent origins in paper mills  \citep{parker2024paper}. Even more concerning, generative %artificial intelligence 
AI tools including those built on large language models (LLMs) can produce highly convincing articles that have the potential to bypass existing detection mechanisms~\citep{majovsky2023artificial, kendall2024risks}, which increases the risk of future retractions.  This indicates that the area of problematic publications is constantly adapting and learning to circumvent existing detectors~\citep{majovsky2023artificial,perkins2023academic}.

Retractions of \emph{biomedical} publications %are
are more pronounced than in other fields~\citep{grieneisen2012comprehensive, bik2016prevalence}, with a recent finding revealing that over a fifth of newly published medical articles have issues that may lead to retraction \citep{sabel2023fake}. Moreover, the problem of retracted articles is not confined to a specific set of offending author institutions. In fact, recent high-profile cases have involved retractions from reputable organizations such as Harvard Medical School \citep{fortune2024harvard}.  Many retractions happen post-publication, often after flawed papers have already influenced research or clinical decisions. Even after retraction, many articles continue to be cited without acknowledgment of their retraction~\cite{vuong2020retractions}. 
This widespread trend poses a serious threat to public health.

These observations highlight the need for an early-warning system that assesses the risk of retraction before peer review, which aims to flag articles that contain methodological flaws, metadata inconsistencies, or other issues of scientific integrity. Identifying these potential retractions can help stop flawed manuscripts early, ease reviewer workloads, and prevent the spread of misinformation.

% Prior attempts addressing this challenge mainly rely on relatively primitive heuristic  techniques~\citep{parker2022experts, shepperd2023analysis,feng2024citation}, 
% including some based on argument quality 
% \citep{freedman2024detecting}. While the threat of retracted articles is likely to grow in the future, the AI community has yet to give this issue the attention it deserves. 
% Three major challenges remain. First, there is no standard benchmark for retraction risk assessment. Second, existing approaches mainly rely on heuristics or simple classifiers, offering limited reasoning ability. Third, metadata such as author reputation and journal information can provide useful context but may also introduce bias.

Three major challenges remain. First, there is no standard benchmark for retraction risk assessment. Second, existing approaches mainly rely on heuristics or simple classifiers, offering limited reasoning ability and explanations. Third, metadata such as author reputation and journal information can improve prediction performance but may also introduce bias. Such bias may lead to unfair assessments of articles from less established researchers or journals. It is therefore important to understand how metadata affects predictions and how to reduce its bias without losing prediction performance.

To address these challenges, our work makes three main contributions.

(1) We introduce \emph{PubMed Retraction}, the first large-scale open-source benchmark for biomedical retraction risk assessment, comprising over 12K real-world articles. The benchmark supports early risk assessment using lightweight information available before peer review, including abstracts and metadata\footnote{Complex features such as tables and images are beyond the scope of the current version of our benchmark.}. It provides a foundation for evaluating both prediction performance and metadata bias.

(2) We propose \texttt{Pub-Guard-LLM}, an open-source LLM-based framework for biomedical retraction risk assessment. The framework predicts retraction risks and provides textual explanations. Our experiments show that \texttt{Pub-Guard-LLM} outperforms various baselines and generates relevant and coherent explanations. We further show that enriched metadata improves prediction performance.

(3) We systematically study metadata bias in LLM-based retraction risk assessment. We first evaluate metadata bias using a framework for transparent bias detection. We then explore several bias mitigation methods, including retrieval-augmented generation (RAG), counterfactual fine-tuning, debate, and argumentation-based approaches. Our results suggest a trade-off between prediction performance and metadata bias, highlighting the challenges of developing fair and reliable retraction risk assessment systems.

We hope that the proposed benchmark, framework, and bias analysis will support future research on reliable and fair AI systems for retraction risk assessment.

\section{Related Work}

The increasing number of retracted scientific articles presents a serious challenge to the research community and society. Previous studies have mainly addressed this problem using rule-based heuristics, such as author affiliations~\citep{sabel2023fake}, citation patterns~\citep{shepperd2023analysis,feng2024citation}, and journal impact factors and author networks~\citep{perez2022threats}. More recently, traditional machine learning methods~\citep{dadkhah2023detection,cabanac2022problematic} and LLM-based approaches~\citep{freedman2024detecting,fletcher2025predicting} have been proposed for detecting retracted articles.

However, despite the growing number of retracted articles, this problem has received limited attention from the AI research community~\citep{byrne2024call}. Three important research gaps remain.

(1) \emph{Benchmarks}.
Existing studies often rely on Retraction Watch~\citep{RetractionWatchDatabase}, a database that collects records of retracted scientific articles, to analyze retraction patterns~\citep{shepperd2023analysis,byrne2022protection}. However, this collection does not provide a standardized benchmark for comparing retraction risk assessment methods. There is still a need for a large-scale benchmark containing both retracted and non-retracted articles.

(2) \emph{Open-Source Tools}.
Although several systems for detecting retracted articles exist, many are proprietary and not publicly available~\citep{christopher2021raw}. While LLM-based approaches have been explored for detecting misinformation~\citep{cao2024can} and plagiarism~\citep{wahle2022large}, open-source LLM-based systems specifically designed for retraction risk assessment remain limited. Such systems could improve transparency and support further research.

(3) \emph{Metadata Bias}.
Existing studies mainly focus on prediction performance, with limited attention to potential bias. Metadata, such as author, institution, and journal reputation, may introduce bias into retraction risk predictions. However, how metadata affects such predictions and how this bias can be mitigated remain largely unexplored.

To address these gaps, we introduce \texttt{Pub-Guard-LLM} and the \emph{PubMed Retraction} benchmark. We evaluate prediction performance and systematically investigate metadata bias and potential mitigation methods.

% \begin{figure*}
%     \centering
%     \scalebox{0.95}{
%     \includegraphics[width=0.88\textwidth]{figures/pub_guard_llm_framework.pdf}}
%     \caption{%An illustration 
%     Framework of the proposed Pub-Guard-LLM (Details in Section~\ref{sec:pubguardLLM})}
%     \label{fig:framework}

% \end{figure*}

\begin{table*}[t] 
        \centering
	\setlength{\tabcolsep}{2.5mm}{
		\begin{threeparttable} 
			\begin{tabular}{c|ccc|cccc}  
				\toprule 
                    &\multicolumn{3}{c|}{General}&\multicolumn{4}{c}{Cancer}\cr
                    Dataset &\textbf{\underline{\texttt{Train}}}&\textbf{\underline{\texttt{Validation}}}& \textbf{\underline{\texttt{Test}}}  &\textbf{\underline{\texttt{Breast}}} &\textbf{\underline{\texttt{Lung}}}
                    &\textbf{\underline{\texttt{Ovarian}}}
                    &\textbf{\underline{\texttt{Colorectal}}}\cr
				\midrule
                    Sample &10,000&500&500&300&300&292&300\cr
                    %Retraction 
                    Retraction Rate & 24.5\% & 25.6\% & 22.4\%& 25.0\%& 38.7\%& 38.0\%&33.0\%\cr
                    High Profile Rate & 34.0\%&  35.9\% &  33.9\%& 38.7\%& 22.4\%& 29.7\%&28.3\%\cr
				\bottomrule
			\end{tabular}
			\caption{Basic statistics of  \emph{PubMed Retraction}.
            High Profile Rate refers to the articles deemed retracted despite their meta-data being ranked high (details in Appendix~\ref{sec:app_benchmark}). 	\label{tab:stat_pubmed_retraction}}
		\end{threeparttable}
	}
	%\end{minipage}%
 \vspace{-10pt}
\end{table*}

\section{\emph{PubMed Retraction} Benchmark% Construction
} 
\label{sec:bench}

In this work, we aim to detect \emph{retraction risk}, which can be defined as the identification and public flagging of a scholarly paper that contains substantial methodological flaws, metadata inconsistencies, or other issues of scientific integrity, which, if substantiated, are likely to lead to retraction.

To address the lack of a standardized benchmark for evaluating and comparing retracted article detection methods, we curate a large open-source benchmark for biomedical articles, \emph{PubMed Retraction}, comprising %nearly 12K\todo{over 11K? we should consistently use one or the other} 
over 12K real-world biomedical articles, including both %legitimate 
non-retracted and retracted publications. We select the biomedical domain %is selected 
due to the higher prevalence of retracted publications therein %, which is 
than in other fields
~\citep{grieneisen2012comprehensive, bik2016prevalence}, as well as the significant %downstream 
impact of this type of retraction in patient safety, clinical decision-making, and public trust in medical research.

To construct \emph{PubMed Retraction}, we use the PubMed database~\citep{wheeler2007database} and retrieve articles using the Unix command-line tool EDirect~\citep{kans2024entrez}. To ensure relevance and avoid outdated data, we focus on articles published between 2000 and 2025. Our goal is to promote the use of minimal textual information for retraction detection, so we design the %dataset 
benchmark to include only %the 
title, abstract, and metadata.
Our benchmark is intended to serve as an early warning system for decision-makers. For example, it can be used before peer review when reviews and citation data are not yet available, which enables journal editors and conference organizers to receive alerts before review assignments are made.
Although we fully acknowledge that including full texts, figures, and tables could enhance this task, we aim to develop a simple yet effective solution based on 
%metadata and abstracts 
minimal information (title, abstract, metadata) in this work.

Specifically, each article in \emph{PubMed Retraction} is structured to include seven key attributes: 
\textit{PubMed ID, Title, Abstract, Authors, Affiliations, Journal, Is
Retracted},
with the latter a binary indicator (Yes or No). 
An article is labeled as %retracted 
\textit{retracted} if the keyword ``retracted'' appears in its publication type attribute. 
To maintain a realistic distribution, we include a lower number of retracted articles than legitimate
ones, which is aligned with real-world cases. %As a result, the final dataset 
Overall, the benchmark consists of 12,192 articles.

Formally, %the dataset
ignoring the partitions, \emph{PubMed Retraction}
can be represented by $\mathcal{D} = \{(x_i, y_i)\}_{i=1}^{N}$, where each $x_i$ is an article (title, abstract and metadata) and $y_i$ is the binary label% to indicate the legitimacy
.
The basic statistics of \emph{PubMed Retraction} %\FT{and its partitions} 
are given in Table~\ref{tab:stat_pubmed_retraction}.

\subsection{Metadata Enrichment}

Besides textual information, metadata provides valuable contextual evidence for assessing retraction risk. Therefore, we enrich each article with metadata collected from publicly available scholarly resources. Unlike previous benchmarks that primarily focus on article text, \emph{PubMed Retraction} includes enriched metadata, which enables future work to study both the performance of metadata-aware methods and the potential bias introduced by metadata.

For this, we identify three  fundamental factors to assess an article's legitimacy: (1) author credibility, (2) affiliation reputation, and (3) journal reputation, and use the relevant external knowledge to enrich the 
input.

\paragraph{Author Credibility.} In order to evaluate the credibility of authors,
we use the Semantic Scholar database~\citep{lo2019s2orc}, which contains millions of articles with extensive metadata and citation information%,
, as an external knowledge source. To quantify author reputation, we use the h-index% as an indicator
. Since LLMs are not inherently sensitive to numerical values, we map h-index scores into five reputation levels
(\textit{very low, low, medium, high, very high})
using a predefined piecewise function (see Appendix~\ref{sec:app_benchmark}), 
e.g., 
 \textit{Geoffrey Hinton (author h-index: 188, %Leading Expert
 very high)}.

\paragraph{Affiliation Reputation.}  As an indicator of the reputation of an author's affiliation, we use the average citation count per institution% is utilized
, determined with %For this, 
OpenAlex~\citep{priem2022openalex} as external knowledge source. This is a fully open index of scholarly works, which provides comprehensive data on the total publications and citations of institutions%, is used as an external knowledge source
. The calculated average citation count is then mapped into five reputation levels (\textit{very low, low, medium, high, very high}), e.g., \textit{Harvard University (institution average citation: 64,
very high).} 

\paragraph{Journal Reputation.} We use the ImpactFactor library\footnote{%https://
impact-factor.readthedocs.io/en/latest/} to obtain the Journal Citation Reports (JCR) partition, which serves as an indicator of a journal’s reputation. To enhance interpretability, we assign a human-readable categorized label (\textit{low, medium, high} or \textit{very high}) to each journal based on its JCR ranking, e.g., \textit{Nature (journal JCR: Q1, 
%Top Level Journal
very high).}

The knowledge retrieved from external resources is integrated into the 
benchmark $\mathcal{D}$ for reducing duplicated queries, 
formally resulting in
$\mathcal{D}' = \{(x_i, k_i, y_i)\}_{i=1}^{N}$ where $k_i$ is the external information for
$(x_i,y_i)\in \mathcal{D}$. 
Note that, if the inquired knowledge is missing from the external knowledge sources, we use, as $k_i$, the keyword \textit{``null''} to indicate the absence. We choose to %include the ``null" keyword in our pipeline, 
do so rather than omitting the field entirely, as we found it does generally indicate a lower article quality and improves the downstream performance.

Since metadata can both improve prediction and introduce undesirable bias, the benchmark provides a metadata-aware and fair retraction risk assessment

\subsection{Data Split}
To evaluate generalization capabilities of models, we partition articles based on keywords to create two subsets as described below. This partitioning strategy is designed for a fair assessment of a model’s ability to handle out-of-distribution samples, a critical aspect for real-world applications where models encounter unseen data.
(1) \emph{General Set}. This set comprises articles covering various diseases and is used for model fine-tuning. It is further split into Train, Validation, and Test subsets. (2) \emph{Cancer Set}. This set consists exclusively of articles related to four types of cancer: Breast, Lung, Ovarian, and Colorectal. Articles of these four categories are %entirely 
excluded from the General Set, which means that %they
these cancers %remain as 
are unseen diseases during fine-tuning and validation. 
Additionally, our benchmark includes high-profile retracted cases, which makes this task particularly challenging (details in Appendix~\ref{sec:app_benchmark}).

\section{\texttt{Pub-Guard-LLM}}
\label{sec:pubguardLLM}

\texttt{Pub-Guard-LLM} is an end-to-end system for retraction risk assessment built upon the \emph{PubMed Retraction} benchmark. Given a biomedical article and its enriched metadata, the system aims to predict its retraction risk while generating a natural-language explanation for the prediction. To achieve this, we first use a strong teacher LLM to generate explanations and then fine-tune a smaller language model to jointly predict retraction labels and produce corresponding explanations. We subsequently study the contribution of enriched metadata to prediction performance and investigate potential metadata-related biases in the resulting predictions.

\subsection{%Knowledge 
Explanation Distillation}

%After augmenting the input, the next step focuses on refining the output to achieve interpretability and improved performance in the model that will be trained on the enhanced input-output pairs. 

The %constructed 
\emph{PubMed Retraction} benchmark %in its initial form 
provides only a binary label (\textit{%retracted
retracted} or \textit{not}) as a target. To construct a model that classifies articles accurately while offering reliable explanations, it is essential to incorporate either human-annotated or machine-generated explanations into the ground truth. Given the high cost of human annotation, we opt for %machine-generated explanations 
the latter
as a more scalable and efficient alternative.

Recent research shows that distilling reasoning capabilities from larger (language) models can significantly enhance the performance of smaller models with fewer learning examples~\citep{li2022explanations, hsieh2023distilling, shridhar2023distilling}: teacher model rationales provide richer insights into why an input is mapped to a specific output label, capturing task-relevant knowledge missing from the %original 
input. What is more interesting is that even if some intermediate steps of the distilled explanations contain inaccuracies, student models can still have enhancements by learning reasoning flows~\cite{li-etal-2023-symbolic,hsieh2023distilling}.
%To leverage this, we 
We use a teacher model to generate explanations for each article via prompting, resulting in
$\mathcal{D}''=\{(x_i, k_i,e_i, y_i)\}_{i=1}^{N}$ where $e_i$ is the corresponding distilled explanations with regard to $(x_i, k_i, y_i)\in \mathcal{D}'$. 
\iffalse This additional distilled knowledge is then %added to the output, which originally consists only of a binary label, to be 
used in the %training 
fine-tuning of 
%our 
a smaller model
to obtain \texttt{Pub-Guard-LLM}.
\fi

%We observe 
Note that instruction-aligned LLMs tend to classify all articles as not retracted, likely to avoid controversial responses. 
This behavior arises because many LLMs, when fine-tuned for alignment and safety, may become overly conservative and ignore contextually relevant information~\citep{rottger2024xstest}. 
To mitigate this issue, it is crucial to assign a clear stance to the teacher model. For example, when evaluating a retracted article, we include the prompt: \textit{``You are tasked with providing explanations and making a firm case for why it has issues and should be retracted. You should have a clear stance''}. This ensures strong, rationale-driven explanations instead of the default neutral or overly cautious responses (see Table~\ref{tab:prompt_explanation} in Appendix for an example prompt). Additionally, we conduct human evaluations to validate the coherence and relevance of distilled explanations (see Section~\ref{sec:eval_explanation} in Appendix).

\subsection{Fine-Tuning}\label{sec:finetuning}
We leverage the samples in \emph{PubMed Retraction} %and 
augmented with the external information and distilled explanations ($\mathcal{D}''$) to fine-tune \texttt{Pub-Guard-LLM}.
\iffalse
Formally, %the dataset
\emph{PubMed Retraction}
can be represented as $\mathcal{D} = \{(x_i, y_i)\}_{i=1}^{N}$, where each $x_i$ is an article (title, abstract and metadata) and $y_i$ is the binary label% to indicate the legitimacy
.
Through the workflow of accessing external knowledge, we %can  add this knowledge into each article
obtain  $(x_i, k_i, y_i)$, where $k_i$ is the external information e.g. drawn from %such as 
the h-index of authors% and JCR ranking of the journal
, as discussed earlier).
%Since binary label learning is not sufficient, 
We then integrate explanations from the teacher model %into this learning process and then the input can be denoted as
to obtain augmented samples $(x_i, k_i, e_i, y_i)$, where $e_i$ is the corresponding generated explanations with regard to $(x_i, k_i, y_i)$.
We apply a curated template to each input for querying the model, but omit its notation here for legibility.
%\todo{I would drop this last sentence - if anything it should go in the previous section}
\fi 
%
To encourage \texttt{Pub-Guard-LLM} to learn both labels and explanations, we introduce a multi-task learning objective. Suppose we have a base model $f_{\theta}$ %, which is
 able to generate label $y$ and explanation $e$ given article $x$ with external information $k$ , i.e., $f_{\theta}(x, k) = (y, e)$. Then, %and
our goal is to optimize $\theta$ for this task.

The model is 
fine-tuned on $y$ to understand how to identify retracted articles via a binary classification task  with%loss the \todo{averaged?} cross entropy between the ground truth and predicted label
:
\begin{equation}
    \mathcal{L}_{\text{cls}} = \frac{1}{N} \sum_{i=1}^{N} \ell \big(f_{\theta}(x_i, k_i), y_i \big)
\end{equation}
where %$\mathcal{L}$
$\ell$ is the cross entropy between the ground truth and the predicted label. To encourage the model to output explanations behind the decision, we introduce a next-token prediction task with% loss the averaged cross entropy between the generated tokens and the distilled explanation from the teacher model
:
\begin{equation}
    \mathcal{L}_{\text{explanation}} = \frac{1}{N} \sum_{i=1}^{N} \ell \big(f_{\theta}(x_i, k_i), e_i \big)
\end{equation}
where %$\mathcal{L}$  
$\ell$ is the averaged cross entropy between the generated tokens and the distilled explanation from %a
the teacher model.

During fine-tuning, %we make 
the model learns to predict both label $y$ %with
and explanation $e$ given an article:
\begin{equation}\label{eq:multi_task}
    \mathcal{L} = \mathcal{L}_{\text{cls}} +\lambda\mathcal{L}_{\text{explanation}}
\end{equation}
where $\lambda$ is a hyper-parameter that controls the balance between the two objectives. 
This learning process helps \texttt{Pub-Guard-LLM} acquire knowledge beyond simple binary labels, which results in a strong performance with well-reasoned explanations.

After fine-tuning, the model $f_\theta$ performs inference on unseen biomedical articles using the article content and enriched metadata. Given an article $x$ and its enriched metadata $k$, the model predicts its retraction risk and generates a textual explanation:
\begin{equation}
(\hat{y}, e) = f_\theta(x, k),
\end{equation}\label{eq:vanilla}
where $\hat{y}$ denotes the predicted retraction label and $e$ is the corresponding explanation.  
We refer to this as the \emph{vanilla} reasoning mode.
% An input-output example is provided in Appendix~\ref{app:expl}.

In addition to vanilla reasoning, we consider RAG and debate as alternative reasoning modes. We investigate their potential for mitigating metadata bias, together with counterfactual fine-tuning and argumentation-based interventions, which is described in Section~\ref{sec:mitigation_approach}.

\begin{table*}[tb] 
	\centering
	\setlength{\tabcolsep}{3.8mm}{
		\begin{threeparttable} 
			\begin{tabular}{cc|c|cccc|c}
                \toprule
                    Model& $|\theta|$ &\texttt{Test}&\texttt{Breast}& \texttt{Lung} &\texttt{Ovarian} & \texttt{Colorectal} &Avg\cr
				\midrule
                    \multicolumn{7}{c}{\textit{Fine-Tuning Binary Classifiers}}\cr
                    \midrule
                    SciBERT  &110M&60.9&56.1&74.0&67.2&71.5&66.0\cr
                    BioLinkBERT  &340M&58.1&57.7&75.4&71.0&72.2&66.9\cr
                    Gatortron  &345M&57.6&58.6&71.2&70.9&67.0&65.1\cr
                    Llama-3.1  &8B&58.1&61.4&73.5&76.6&61.3&66.8\cr
                    Bio-Llama  &8B&59.1&54.8&70.5&71.7&66.9&64.6\cr
                    %Llama-3.1  &8B&60.9&63.2&76.5&72.3&68.3&68.2\cr
                    %Bio-Llama  &8B&56.4&61.7&63.9&66.7&61.6&62.1\cr
                    %BioLinkBERT + PEARL  &210M&67.6&69.6&79.0&70.5&69.8&71.3\cr
                    %Gatortron + PEARL &445M&66.3&69.3&78.3&76.2&70.8&72.2\cr
                     \midrule
                    \multicolumn{7}{c}{\textit{Zero-Shot Reasoning}}\cr
                    \midrule
                    Llama-3.1-Instruct  &8B&29.8&20.0&46.2&43.3&43.6&36.6\cr
                    OpenScholar  &8B&11.5&18.8&20.3&19.9&14.1&16.9\cr
                    Bio-Llama  &8B&37.7&40.7&56.7&59.3&47.9&48.5\cr
                    PMC-Llama  &13B&36.0&36.4&56.7&23.3&25.8&35.6\cr
                    \midrule
                    \multicolumn{7}{c}{\textit{Ours}}\cr
                    \midrule
                    %Pub-Guard-LLM &8B &\textbf{61.2}&\textbf{68.4}&\textbf{78.3}&74.8&\textbf{72.9}&\textbf{71.1}\cr
                    % \multicolumn{7}{c}{\textit{Ours}}\cr
                    % \midrule
                    Pub-Guard-LLM & 8B & \textbf{68.6} & \textbf{76.7} & \textbf{82.3} & \textbf{77.4} & \textbf{75.8} & \textbf{76.2} \cr
                    % Pub-Guard-LLM (\textit{Debate}) &8B\cr 
				\bottomrule
			\end{tabular}
			\caption{Performance on the \emph{PubMed Retraction} benchmark. We report F1 scores on the General Test set and four out-of-distribution cancer subsets.} \label{tab:overall}
		\end{threeparttable}
	}
	%\end{minipage}%
	\vspace{-10pt}
\end{table*}

\section{Experimental Set-Up}

\subsection{Baselines}
We compare our approach to the following% baselines
:

(1) \emph{Fine-tuning Binary Classifiers}. This method appends a classification head—a linear layer—on top of a language model
 to handle retraction detection% tasks
.  However, this method can only output binary labels (no explanations). The title, abstract, and metadata are concatenated using a special token, [SEP], and this combined textual input is fed into the classifier.  %The following language models are used  for comparison
Specifically, we use language models: SciBERT~\citep{beltagy2019scibert}, BioLinkBERT~\citep{yasunaga2022linkbert}, Gatortron~\citep{yang2022gatortron}, Llama-3.1~\citep{dubey2024llama} and Bio-Llama~\citep{ContactDoctor_Bio-Medical-Llama-3-8B}.
We also find that introducing additional models to represent metadata separately can be beneficial. Thus, we apply a lightweight short-text model, PEARL~\citep{chen2024learning}, to represent metadata and obtain embeddings for authors, affiliations, and journal. We continue to use the original language model to represent the title and abstract. Subsequently, these embeddings are concatenated and fed into the classification head. We denote %this
the resulting variant of language model \textit{X}  as \textit{``X + PEARL''}.

(2) \emph{Zero-shot Reasoning}. 
%LLMs are generally known as few-shot reasoners, meaning they can perform powerful reasoning with few learning examples~\citep{brown2020language, kojima2022large}. 
Compared to binary classifiers, LLMs can offer answers with explanations by using prompt-based methods~\cite{kojima2022large}.
We provide LLMs with a dedicated prompt to obtain binary answers with explanations. Here, we use the following instruction-aligned LLMs:
Llama-3.1-Instruct~\citep{dubey2024llama}, OpenScholar~\citep{asai2024openscholar}, Bio-Llama~\citep{ContactDoctor_Bio-Medical-Llama-3-8B} and PMC-Llama~\citep{wu2024pmc}.

\subsection{Implementation Details}
All approaches are implemented with PyTorch \citep{paszke2019pytorch} and HuggingFace~\citep{wolf2020transformers}. 
We use Amazon EC2 \textit{g5.12xlarge} instances with 4$\times$24 GiB for all experiments.
The teacher model used for distilling explanations is Mistral-Large~\citep{mistral_large_2024}. Our model fine-tunes Llama-3.1-8B~\citep{dubey2024llama} using LoRA~\citep{hulora} adapters (r=128, lora\_alpha=128, lora\_dropout=0.1). We train on the train set in the \textit{General} partition for one epoch and AdamW 8-bit optimizer with a learning rate of 1e-4, batch size 4, gradient accumulation of 4 and the $\lambda$ is 1.  

All baseline binary classifiers are fine-tuned %using 
with three different seeds: we report their performance based on the best validation set score.

\section{Results}

\subsection{Performance of Pub-Guard-LLM}
\label{sec:overall}
Table~\ref{tab:overall} reports the performance of different approaches on the \emph{PubMed Retraction} benchmark, including the general test set and four out-of-distribution (OOD) cancer subsets. 

Overall, \texttt{Pub-Guard-LLM} consistently outperforms all baseline methods across all five evaluation sets, achieving an average F1 score of 76.2\%. Compared with BioLinkBERT, the strongest baseline with an average F1 score of 66.9\%, our approach achieves an improvement of 9.3 percentage points. Notably, \texttt{Pub-Guard-LLM} achieves the highest F1 score on each OOD subset, including Breast (76.7\%), Lung (82.3\%), Ovarian (77.4\%), and Colorectal (75.8\%). These results demonstrate the effectiveness of \texttt{Pub-Guard-LLM} for retraction risk assessment and its ability to generalize across different biomedical domains.

Among the baseline methods, fine-tuned binary classifiers consistently outperform zero-shot reasoning approaches. In particular, BERT-based models achieve competitive performance despite their relatively small parameter sizes. For example, BioLinkBERT (340M parameters) achieves an average F1 score of 66.9\%, slightly outperforming the fine-tuned Llama-3.1 model (8B parameters), which achieves 66.8\%. These findings suggest that larger model size does not necessarily translate into better performance under conventional classification fine-tuning, consistent with previous findings~\citep{lehman2023we, chen2024role}.

In summary, \texttt{Pub-Guard-LLM} achieves the best overall performance and consistently outperforms existing approaches across both the general test set and OOD cancer subsets. These findings highlight its potential for retraction risk assessment across biomedical research domains. Additionally, we conduct an ablation study to examine the contributions of key components, including the \textit{Base LLM}, \textit{Teacher Model}, and \textit{External Knowledge}. The results are presented in Appendix Table~\ref{tab:ablation}.

\begin{figure}
    \centering
    \includegraphics[width=0.65\textwidth]{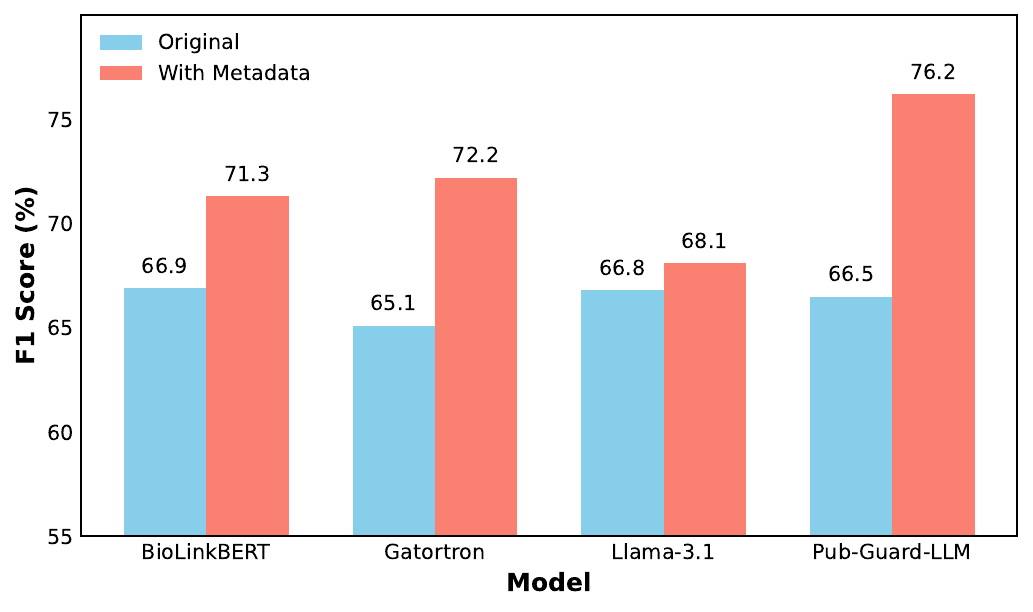}

    \caption{Effect of enriched metadata on retraction prediction performance across different models. Bars show the average F1 scores (\%) with and without enriched metadata.
    }
    \label{fig:metadata}
\end{figure}

\subsection{Effect of Enriched Metadata}\label{sec:effect_metadata}
Although enriched metadata provides valuable contextual evidence, it remains unclear whether it consistently improves retraction risk assessment. We therefore compare models with and without enriched metadata. Figure~\ref{fig:metadata} summarizes the results. Enriched metadata consistently improves all evaluated models, with gains ranging from 1.3 to 9.7 F1 points. \texttt{Pub-Guard-LLM} also benefits substantially, improving from 66.5 to 76.2 F1. These results demonstrate that enriched metadata provides useful signals beyond article title and abstract.

% \begin{table}[t]
% \centering
% \scriptsize
% \label{tab:bias_integration_accuracy}
% \begin{tabular}{lcccc}
% \toprule
% \textbf{Semantics}
% & \textbf{Original}
% & \textbf{QBAF}
% & \textbf{QBAF + Score}
% & \textbf{QBAF + Graph} \\
% \midrule
% QuAD
% & 82.12
% & 82.03
% & 82.03
% & 82.07 \\

% DF-QuAD
% & 82.12
% & 78.70
% & 80.34
% & 79.74 \\

% QuEM
% & 82.12
% & 80.43
% & 81.39
% & 80.38 \\

% MLP
% & 82.12
% & 81.16
% & 81.75
% & 81.16 \\
% \bottomrule
% \end{tabular}
% \caption{Prediction accuracy under different strategies.}
% \end{table}

% \begin{table}[t]
% \centering
% \small
% \label{tab:bias_mitigation}
% \begin{tabular}{lccc}
% \toprule
% \textbf{Method} & \textbf{Before (\%)} & \textbf{After (\%)} & $\boldsymbol{\Delta}$  \\
% \midrule
% Graph-QuAD    & 50.05 & 50.00 & -0.05 \\
% Graph-DF-QuAD & 53.42 & 53.28 & -0.14 \\
% Graph-QuEM    & 51.32 & 51.23 & -0.09 \\
% Graph-MLP     & 50.96 & 50.96 &  0.00 \\
% \midrule
% Score-QuAD    & 50.05 & 48.86 & -1.19 \\
% Score-DF-QuAD & 53.42 & 52.69 & -0.73 \\
% Score-QuEM    & 51.32 & 51.19 & -0.14 \\
% Score-MLP     & 50.96 & 50.55 & -0.41 \\
% \midrule
% Fine-tune     & --    & 49.95 & -- \\
% \bottomrule
% \end{tabular}
% \caption{Bias rate before and after mitigation.}
% \end{table}

% \begin{figure*}
%     \centering
%     \includegraphics[width=0.95\linewidth]{figures/best_method_bias_rate_barplot.png}
%     \caption{Caption}
%     \label{fig:placeholder}
% \end{figure*}

\begin{table}[t]
\centering
\begin{tabular}{llrr}
\toprule
\textbf{Metadata} & \textbf{Reputation Level} & \textbf{Number} & \textbf{Ratio (\%)} \\
\midrule
\multirow{2}{*}{Author}
    & High & 12 & 0.5 \\
    & Low  & 166 & 7.6 \\
\midrule
\multirow{2}{*}{Affiliation}
    & High & 18 & 0.8 \\
    & Low  & 282 & 12.9 \\
\midrule
\multirow{2}{*}{Journal}
    & High & 0 & 0.0 \\
    & Low  & 1011 & 46.1 \\
\bottomrule
\end{tabular}
\caption{Metadata bias assessment of Pub-Guard-LLM using ABIDE on 2,192 test samples. Number denotes the count of detected bias cases for each metadata category and reputation level, while Ratio represents the percentage relative to all evaluation samples.}
\label{tab:metadata_bias}
\end{table}

\subsection{Metadata Bias Assessment}

Although enriched metadata improves prediction performance, it may also encourage models to over-rely on reputation-related signals. We therefore investigate metadata bias in \texttt{Pub-Guard-LLM} using ABIDE (Argumentative Bias Detection)~\citep{ayoobi2026argumentative}, a model-agnostic framework for transparent bias detection.
Specifically, we examine whether metadata influences the model's predicted binary labels. Bias in generated explanations is discussed separately in Section~\ref{sec:expl}.

ABIDE~\citep{ayoobi2026argumentative} is a transparent bias detection framework that identifies whether similar samples receive different predictions across groups. 
In our setting, we use Pearl~\citep{chen2024learning} embeddings of article titles and abstracts to represent their semantic content. Based on these embeddings, we construct local neighborhoods of semantically similar articles. 
Within each neighborhood, articles are divided into high- and low-reputation groups according to their author, affiliation, or journal metadata. ABIDE then compares the model's prediction outcomes between these groups to identify potential bias. It uses Quantitative Bipolar Argumentation Frameworks (QBAFs) to combine evidence of group advantage and disadvantage, while considering the significance, diversity, and objectivity of each neighborhood. This argumentation-based approach makes the bias assessment transparent and interpretable. 
%Further implementation details are provided in the appendix.

Table~\ref{tab:metadata_bias} reports the number and percentage of detected bias cases across 2,192 samples from the \emph{PubMed Retraction} benchmark, excluding the training set. The results show that bias is more frequently detected for low-reputation groups across all metadata types, particularly for journal reputation, where 46.1\% of samples have detected bias. These findings suggest that \texttt{Pub-Guard-LLM} may exhibit reputation-related bias, which motivates further investigation into bias mitigation.

\section{Bias Mitigation}
Our previous analysis shows that enriched metadata can introduce bias into retraction risk assessment. To address this issue, we investigate four potential mitigation strategies: Retrieval-Augmented Generation (RAG), Debate, Counterfactual Fine-tuning, and Argumentation-based Debiasing. These approaches aim to reduce the model's reliance on metadata from different perspectives. We then test whether these strategies can reduce metadata bias while maintaining prediction performance.

\subsection{Bias Mitigation Approaches}~\label{sec:mitigation_approach}

\paragraph{\textbf{Retrieval-Augmented Generation (RAG)}} In the vanilla reasoning mode, $f_{\theta}(x, k)$ (see Section~\ref{eq:vanilla}) does not store all articles within %their
its parameters, and, in some cases, an input article’s main argument may lack support from existing publications. This suggests that the article's findings could be controversial and potentially retracted. To address this, we retrieve relevant top-$l$ articles $\mathcal{A}$ (\textit{all legitimate}) to the input article $x$ from PubMed and use them to validate the claims made in $x$
by passing them as input into the model: $f_{\theta}(x, k, \mathcal{A})$.

\begin{figure}[t]
    \centering
    \includegraphics[width=0.75\textwidth]{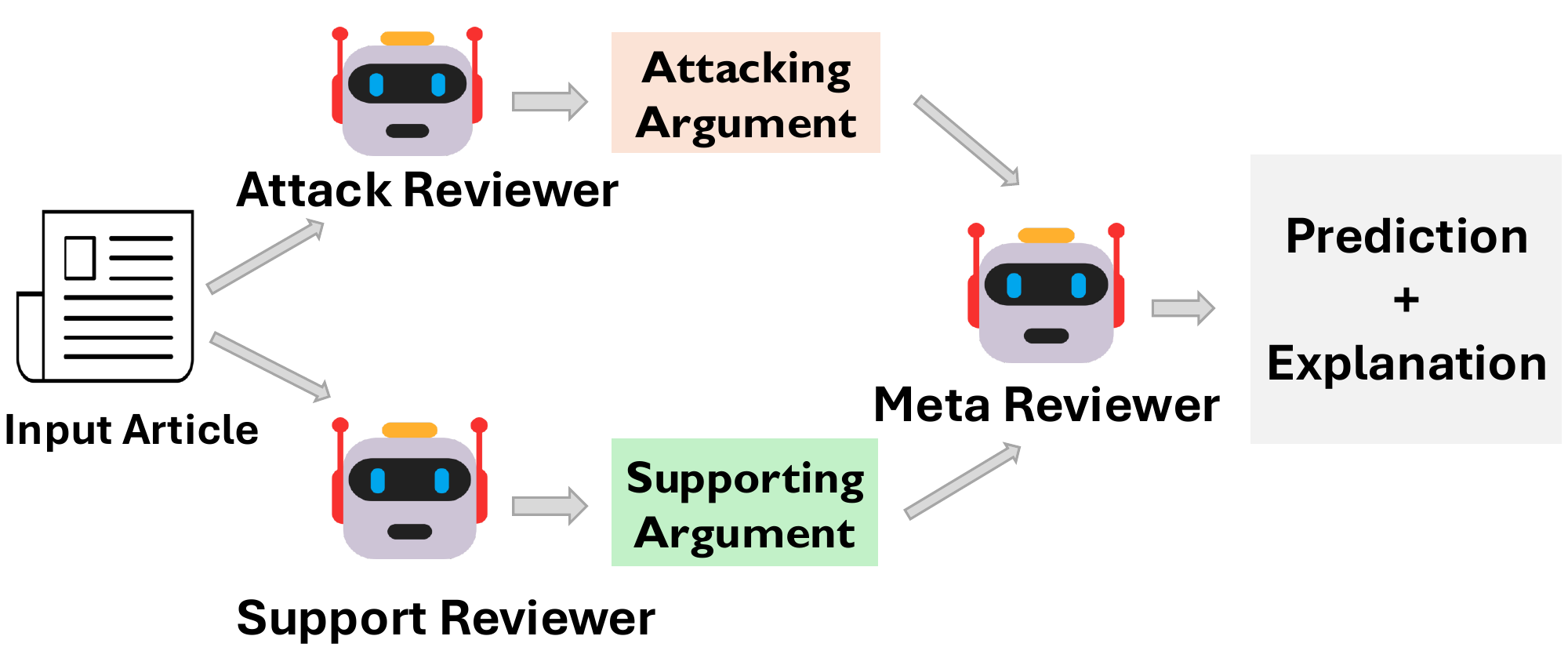}

    \caption{Given an input article, the Attack and Support Reviewers generate opposing arguments regarding its retraction risk. The Meta Reviewer then evaluates both perspectives to produce a final prediction and a natural-language explanation.
    }
    \label{fig:debate_framework}
\end{figure}

\paragraph{\textbf{Debate}} Recent research has shown that Multi-Agent Debate can enhance both factual accuracy and explainability in LLMs~\citep{liang2023encouraging, duimproving, freedman2024argumentative}. The core idea involves multiple agents presenting their arguments, while a judge oversees %the debate process 
to reach a final decision. 

To generate the debate process in our setting (Figure~\ref{fig:debate_framework}), we fine-tune three specialized \texttt{Pub-Guard-LLM}s following the steps in Section~\ref{sec:finetuning}:
(1) a support reviewer $f_{\theta}^{s%upport
}$, trained on legitimate articles, generates supporting arguments $arg^{+}$;
(2) an attack reviewer $f_{\theta}^{a%ttack
}$, trained on retracted articles, generates attacking arguments $arg^{-}$.
Thus, given input article $(x, k)$, the support reviewer $f_{\theta}^{s%upport
}$ consistently provides arguments in favor of the article while the attack reviewer $f_{\theta}^{a%ttack
}$ presents counterarguments highlighting potential issues.
Following this debate,  we introduce (3) a meta-reviewer $f_{\theta}^{meta}$ that is responsible for making the final decision based on the arguments presented by both reviewers.
To train the meta-reviewer, we use a teacher model to generate debate-based explanations $\hat{e}$, which are then used to fine-tune $f_{\theta}^{meta}$ by reusing the multi-task learning in Equation~\ref{eq:multi_task}. %Finally, 
Overall, the debate-based detection framework is formulated as: $f_{\theta}^{meta}(x, k, arg^{+}, arg^{-})$, where the meta-reviewer makes a decision by considering both the supporting and attacking arguments.

\paragraph{\textbf{Counterfactual Fine-tuning}}
To mitigate the model's reliance on reputation-related metadata, we adopt a counterfactual data augmentation strategy. 
Specifically, we construct counterfactual training examples by modifying one metadata (author, affiliation, or journal) at a time, while keeping the article's title, abstract, and ground-truth label unchanged. 
For author reputation, we modify the H-index and corresponding reputation description. 
For affiliation reputation, we adjust the institution's average citation score, 
and for journal reputation, we change the journal's JCR quartile. These modifications reverse the reputation level from high to low or vice versa, while preserving the original entity names and all other metadata attributes.

For each training sample with available reputation metadata, we randomly select one dimension and generate an additional counterfactual example, which is combined with the original training data. We then continue fine-tuning \texttt{Pub-Guard-LLM} using LoRA, supervising the retraction prediction labels rather than the generated explanations. By exposing the model to articles with identical scientific content and labels but contrasting reputation signals, this approach encourages prediction consistency across different metadata conditions and reduces reliance on potentially biased reputation cues.

\paragraph{\textbf{Argumentation-based Debiasing}}
We explore argumentation-based debiasing to mitigate reputation-related metadata bias in retraction risk predictions. Specifically, we use the fine-tuned Debate mode of \texttt{Pub-Guard-LLM} to generate supporting and attacking arguments, which are organized into a Quantitative Bipolar Argumentation Framework (QBAF)~\citep{baroni2018how}. This framework enables transparent evidence aggregation and targeted interventions on potentially biased arguments.

We construct a one-layer QBAF in which all supporting and attacking arguments are directly connected to the target prediction node. Each argument is assigned an initial weight using a Natural Language Inference (NLI) model\footnote{\texttt{cross-encoder/nli-deberta-v3-large}} to estimate its evidential support from the input article. We also identify arguments associated with reputation-related metadata, including author, affiliation, and journal information. Using the metadata bias evidence identified by ABIDE, we investigate two intervention strategies that modify either argument weights or the argumentation graph.

\textbf{Score Intervention.}
This strategy reduces the influence of metadata-related arguments according to their detected bias levels. For each article, we associate ABIDE's bias assessment with the corresponding metadata features and assign a bias strength $b_i \in [0,1]$ to each relevant argument. If an argument is associated with multiple metadata features, their bias scores are combined using the maximum value. We then adjust the argument's initial weight as follows:
\begin{equation}
w_i^{\mathrm{mitigated}}
=
w_i^{\mathrm{init}}(1-b_i)^{\beta},
\end{equation}
where $w_i^{\mathrm{init}}$ is the initial argument weight and $\beta \geq 0$ controls the intervention strength. A higher bias score results in stronger attenuation, while arguments without detected metadata bias retain their original weights. The same adjustment is applied to both supporting and attacking arguments to avoid favoring either prediction direction. We evaluate different values of $\beta$ and select the one achieving the highest average prediction accuracy.

\textbf{Graph Intervention.}
Rather than reducing argument weights directly, this strategy incorporates the Bias Argumentation Graph (BAG) generated by ABIDE into the original QBAF. For each article, we identify the BAG nodes representing the aggregated bias assessments for author, affiliation, and journal reputation. 
We then extract the relevant subgraphs, preserving their internal supporting and attacking relations, and connect their aggregated bias nodes to the corresponding metadata-related arguments in the original QBAF through attacking edges. Consequently, the influence of metadata-related arguments is reduced through the argumentation structure itself, rather than through an explicit weight adjustment. This allows the final prediction to reflect both the original supporting and attacking evidence and the structured evidence of potential metadata bias.

For both intervention strategies, we adopt QuAD~\citep{baroni2015automatic} semantics to aggregate the strengths of supporting and attacking arguments and compute the final prediction score. The score is converted into a binary retraction prediction using a threshold of $0.5$. We evaluate the effectiveness of both interventions in terms of predictive performance and metadata bias reduction.

\subsection{Mitigation Performance}
% Table~\ref{tab:bias_mitigation} presents the results of different metadata bias mitigation strategies. Overall, the results reveal a trade-off between prediction performance and bias mitigation. RAG achieves the largest reduction in total bias cases, from 1,489 to 276 (81.5\%), but substantially reduces the F1 score. In contrast, Debate achieves a more balanced outcome, reducing total bias cases by 30.5\% while maintaining a relatively high F1 score of 73.5. Argumentation-based methods also reduce metadata bias, but their improvements are smaller and come with greater performance degradation. These findings suggest that Debate offers the most promising balance between prediction performance and metadata bias mitigation among the methods.

Table~\ref{tab:bias_mitigation} compares the prediction performance and metadata bias of different mitigation strategies. Overall, the results reveal a trade-off between prediction performance and the reduction of ABIDE-detected bias cases. RAG achieves the largest reduction in total bias cases (81.5\%), but substantially decreases the F1 score. In contrast, Debate achieves a more favorable balance, which reduces total bias cases by 30.5\% while maintaining an F1 score of 73.5.

However, the effectiveness of bias mitigation varies across metadata attributes. In particular, Debate substantially reduces detected bias cases associated with low journal reputation, while increasing those associated with low author and affiliation reputation. This suggests that reducing overall metadata-related disparities does not necessarily improve outcomes across all reputation groups.

Counterfactual fine-tuning and argumentation-based interventions also reduce the total number of detected bias cases, but at the cost of lower prediction performance. These findings highlight the difficulty of mitigating metadata-related bias without compromising predictive accuracy. 
Among the evaluated methods, Debate provides the most favorable overall trade-off.

\begin{table*}[t]
\centering
\resizebox{\textwidth}{!}{
\begin{tabular}{lcccccccc}
\toprule
\multirow{2}{*}{Method}
& \multirow{2}{*}{F1 ($\uparrow$)}
& \multirow{2}{*}{Total Bias Cases ($\downarrow$)}
& \multicolumn{2}{c}{Author}
& \multicolumn{2}{c}{Affiliation}
& \multicolumn{2}{c}{Journal} \\
\cmidrule(lr){4-5}
\cmidrule(lr){6-7}
\cmidrule(lr){8-9}
& & & High & Low & High & Low & High & Low \\
\midrule
Pub-Guard-LLM (Vanilla)
& 76.2 & 1489
& 12 & 166 & 18 & 282 & 0 & 1011 \\
\midrule
RAG
& 49.9 & 276
& 55 & 28 & 20 & 41 & 18 & 114 \\
Debate
& 73.5 & 1035
& 5 & 198 & 1 & 348 & 2 & 481 \\
Counterfactual Fine-tuning & 69.0 & 1070 & 3 & 231 & 24 & 270 & 0 & 542 \\
Argumentation (Score)
& 70.7 & 1168
& 6 & 191 & 3 & 416 & 1 & 551 \\
Argumentation (Graph)
& 71.3 & 1173
& 6 & 186 & 3 & 415 & 1 & 562 \\
\bottomrule
\end{tabular}
}
\caption{Prediction performance and metadata bias across different
mitigation methods. Total bias cases sum the detected cases across
six metadata reputation groups. The argumentation semantics here is QuAD. }
\label{tab:bias_mitigation}
\end{table*}

\begin{figure*}[t]%
	\centering
	\subfloat[\centering Explanation reliability using LLM-as-a-Judge (average over three runs)]{{\includegraphics[width=0.45\textwidth]{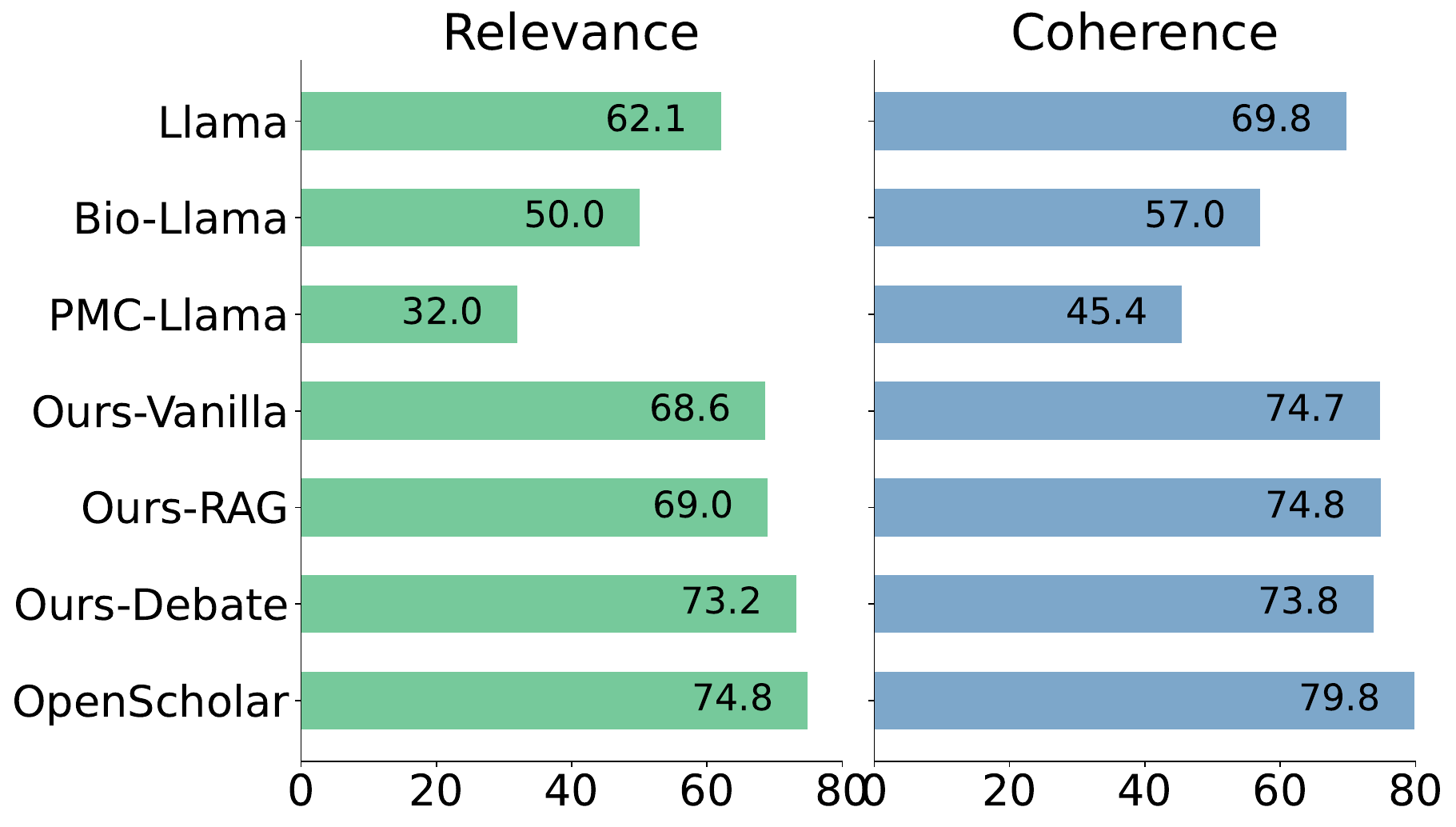} }~\label{fig:eval_explanation}}%
	%\qquad
	\subfloat[\centering Comparison of explanation reliability and label-explanation agreement in the user study (see details in Appendix~\ref{sec:app_user_study})]{{\includegraphics[width=0.45\textwidth]{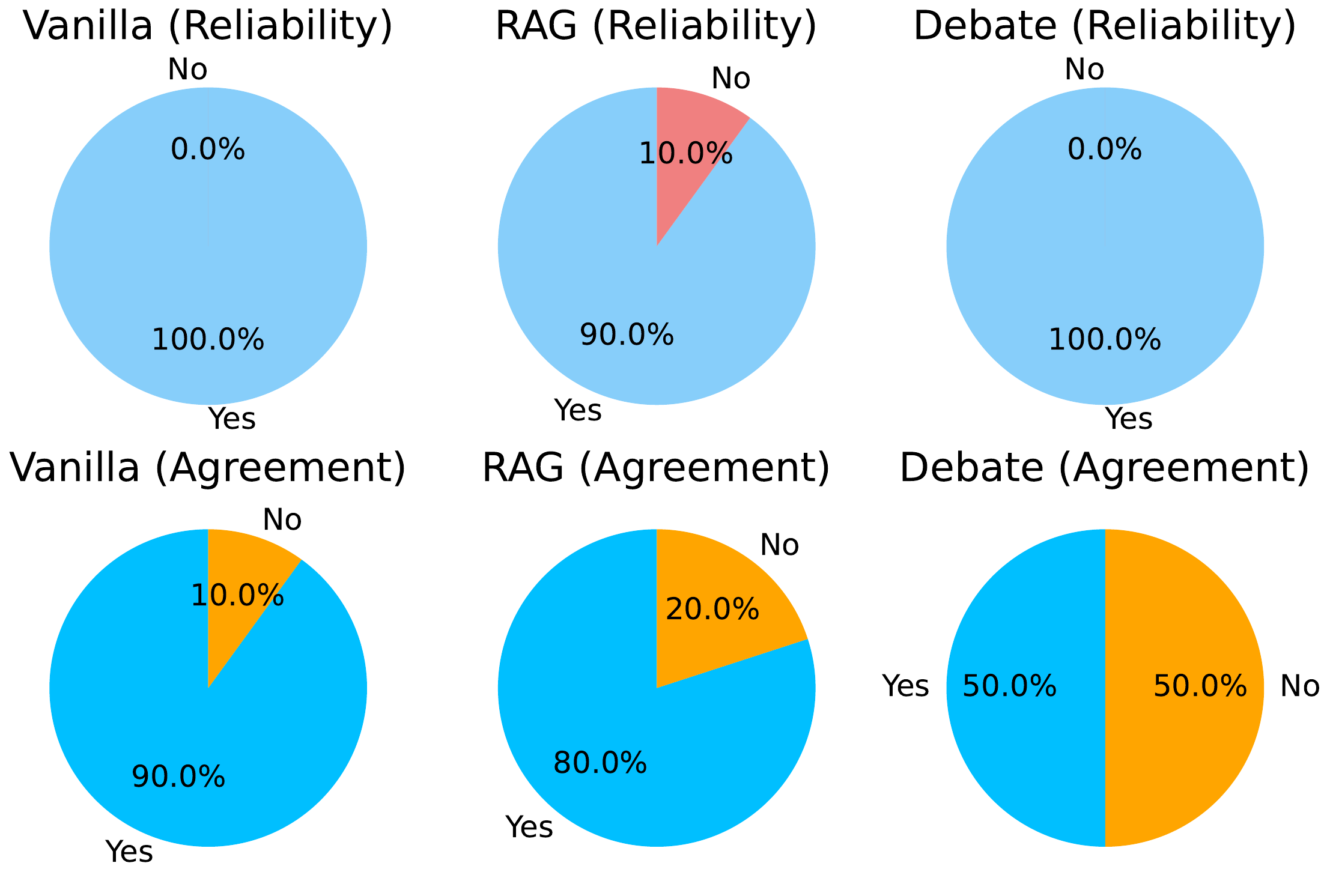} }~\label{fig:user_study}}%
	\caption{Explanation evaluation and user study}
	\label{fig:two_eval_explanation}%
    \vspace{-10pt}
\end{figure*}

\section{Explanation Evaluation}
\label{sec:expl}
Our previous experiments evaluate \texttt{Pub-Guard-LLM} in terms of prediction performance and metadata bias, which solely focuses on the predicted binary labels. However, an important advantage of LLM-based retraction risk assessment is the ability to provide textual explanations. These explanations can help users understand the model's predictions, but they may also introduce additional concerns about reliability and bias. We therefore evaluate explanation quality through both automatic evaluation and a user study, and discuss the remaining challenges of bias in generated explanations.

\subsection{Automatic Evaluation}
Compared to BERT-based models, a distinct benefit of \texttt{Pub-Guard-LLM} is the %reliable 
explanations. 
To evaluate the reliability of LLM-generated explanations, we consider two dimensions introduced, for different settings, in prior work~\citep{kotonya2020explainable,kotonya2024towards, valentino2024introductory}.
(1) \emph{Relevance}. A relevant explanation should highlight the particular features and patterns within the article that the model identified as indicative of retraction or not. The model should only reference information that is present in the article, avoiding hallucinations. (2) \emph{Coherence}. 
This refers to the logical consistency and clarity of the explanation provided by the model for its predictions. Coherence ensures that the explanation flows logically, %and the narrative is easy to follow, 
thereby enhancing the user's understanding of why an article is classified as retracted or not. 
To measure these properties, we adopt the LLM-as-a-Judge paradigm, which has been shown to align with both controlled and crowdsourced human preferences when using strong LLM judges like GPT-4 ~\citep{zheng2023judging}.
We design prompts (see Appendices \ref{tab:relevance_prompt} and 
\ref{tab:coherence_prompt}) for these two dimensions and ask GPT-4o to assess the provided explanations, which assigns a score between 1 and 10.  
The results, presented in Figure~\ref{fig:eval_explanation}, show that our explanations outperform those of other LLMs, with the exception of OpenScholar.
While OpenScholar demonstrates strong explanation quality, its retraction detection capability is significantly lower than other LLMs, with an average F1 score of just 16.9 (see Table~\ref{tab:overall}).
%Additionally, 
Also, note that the debate mode can reduce %hallucinated
irrelevant explanations by achieving higher relevance scores than the another two modes.
These findings validate that \texttt{Pub-Guard-LLM} not only achieves powerful performance but also provides reliable explanations.
However, these quality metrics do not directly measure whether the explanations contain bias.

\subsection{User Study}
\label{sec:user}
We conducted a pilot user study to evaluate whether the outputs of \texttt{Pub-Guard-LLM} align with users' needs. To achieve this, we designed questionnaires covering our three application modes and gathered feedback from three experienced doctors (see details in Appendix~\ref{sec:app_user_study}).
The questionnaire focused on two key aspects: (1) \textit{Reliability:} is the explanation coherent and relevant?  (2) \textit{Agreement:} do users %you
agree with the predicted label and the explanation?
% \begin{wrapfigure}{r}{7.5cm}
%     \centering
%     \includegraphics[width=0.5\textwidth]{figures/user_study.pdf}

%     \caption{Comparison of explanation reliability and label-explanation agreement in the user study (see details in Appendix~\ref{sec:app_user_study}).
%     }
%     \label{fig:user_study}
% \end{wrapfigure}
Figure~\ref{fig:user_study} presents the user study results. The feedback suggests that doctors find our model’s explanations reliable, but there is a discrepancy between expert opinions and machine-generated predictions regarding the debate mode. However, due to the limited number of participants, it is difficult to determine which application mode performs better.

These findings also highlight a limitation of our current approach. While our bias assessment and mitigation strategies focus on the predicted labels, metadata bias may also appear in the generated explanations (see details in Section~\ref{sec:app_user_study}). Future work should therefore investigate how to evaluate and mitigate bias in explanation.

\section{Conclusion}
In this work, we introduce a publicly available benchmark and the first LLM-based system for biomedical retraction risk assessment. Our \texttt{Pub-Guard-LLM} outperforms existing baselines. Beyond prediction performance, we investigate metadata-related bias and evaluate several mitigation strategies. 
Our findings reveal a trade-off between prediction performance and metadata bias mitigation, with Debate achieving the most promising balance. This highlights the need for future research to develop more effective debiasing strategies that preserve prediction performance while reducing reliance on metadata.

Our results also highlight several challenges and directions for future research. First, publication reputation should not be treated as a reliable indicator of scientific integrity, as retracted articles may also appear in prestigious journals. Second, reliance on metadata may introduce bias against junior researchers and emerging institutions. Although we investigate mitigation strategies for prediction-level bias, our user study suggests that metadata bias may also persist in  explanations. Future work should therefore explore methods to detect and mitigate explanation-level bias. Finally, assessing scientific integrity requires evidence beyond metadata and abstracts. Incorporating full texts, tables, and figures may help models better evaluate scientific content.

\section*{Limitations}

Our study has several limitations.

\textbf{(1) Potential catastrophic forgetting.}
\texttt{Pub-Guard-LLM} is fine-tuned for retraction risk prediction. This task-specific training may reduce its ability to perform general reasoning tasks, a problem known as catastrophic forgetting~\citep{mccloskey1989catastrophic}. We do not evaluate its general reasoning ability in this study.

\textbf{(2) Small user study.}
Our user study involves only three participants, which limits the generalizability of its findings. A larger study is needed to better evaluate the generated explanations and compare the three application modes.

\textbf{(3) Remaining metadata bias.}
Our bias evaluation shows that \texttt{Pub-Guard-LLM} can rely on reputation-related metadata, including author, affiliation, and journal information. Although we explore several methods to reduce these biases, none completely removes them. Moreover, our evaluation focuses on selected reputation-related features, and other forms of bias may remain unexplored.

\textbf{(4) Binary classification of retraction status.}
Our dataset uses only two labels: retracted and not retracted. However, papers may be retracted for different reasons, including scientific misconduct, ethical issues, and honest errors. The current model does not distinguish between these reasons. Future work should consider more detailed labels to provide a better understanding of retraction risk.

\section*{Ethics and Privacy Statement}

This study uses the PubMed Retraction benchmark, which contains publicly available information about scientific publications, including author names, affiliations, and retraction status. Although these data are publicly accessible, their use in automated retraction risk assessment raises privacy and reputational concerns. In particular, incorrect predictions may unfairly damage the reputations of researchers and institutions.

A key ethical concern is the potential for metadata-related bias. Models that rely on author reputation, institutional prestige, or journal rankings may reproduce existing inequalities in academia, potentially disadvantaging early-career researchers, emerging institutions, and less-established journals. 
We systematically investigate these biases and evaluate several mitigation strategies. However, our findings indicate that reducing bias often comes at the cost of predictive performance, and residual biases may remain. Moreover, our bias assessments identify potential disparities rather than establishing definitive evidence of discrimination.

Importantly, paper retraction does not necessarily indicate scientific misconduct, as retractions may result from honest errors or other circumstances. Similarly, a predicted retraction risk does not establish that a publication is unreliable or fraudulent. Therefore, \texttt{Pub-Guard-LLM} should not be used to rank researchers or institutions, make accusations of scientific misconduct, or automatically reject manuscripts. Its predictions and explanations should serve only as supplementary information to support careful human review, rather than as definitive judgments about scientific integrity.

Finally, we conducted a small-scale user study involving three clinicians to evaluate the reliability of model-generated explanations and their agreement with predicted labels. The evaluation was based on published article information.

% \section*{Acknowledegments}
% This research was partially supported by the UK Research and Innovation (UKRI) INDICATE project (Grant No. EP/Y017749/1) and the European Research Council (ERC) under the European Union's Horizon 2020 research and innovation programme (Grant Agreement No. 101020934, ADIX).

\section*{Generative AI Usage Statement}
The authors used OpenAI's GPT-6 to assist with language polishing and experimental code development and debugging. Generative AI was not used to independently formulate research hypotheses or draw scientific conclusions. All AI-assisted content and code were reviewed.

%%
%% The next two lines define the bibliography style to be used, and
%% the bibliography file.
\bibliographystyle{ACM-Reference-Format}
\bibliography{sample-base}

%%
%% If your work has an appendix, this is the place to put it.

\appendix

\clearpage
% \newpage
\appendix
\setcounter{table}{0}   
\setcounter{figure}{0}
\renewcommand{\thetable}{A\arabic{table}}
\renewcommand{\thefigure}{A\arabic{figure}}
\setcounter{equation}{0}
\setcounter{subsection}{0}
\renewcommand{\theequation}{A.\arabic{equation}}

\section{Details of Benchmark and External Knowledge}
\label{sec:app_benchmark}
\paragraph{Author Credibility}
To categorize authors based on their reputation, we define a piecewise function that maps h-index values into discrete reputation levels. This approach ensures that LLMs can interpret author credibility effectively by using textual labels instead of raw numerical values. The categorization is as follows:
\begin{itemize}
    \item Emerging Researcher (very low): \( 0 \leq h\text{-index} \leq 5 \)
    \item Early Career Researcher (low): \( 6 \leq h\text{-index} \leq 15 \)
    \item Established Researcher (medium): \( 16 \leq h\text{-index} \leq 30 \)
    \item Influential Researcher (high): \( 31 \leq h\text{-index} \leq 45 \)
    \item Leading Expert (very high): \( h\text{-index} > 45 \)
\end{itemize}

\paragraph{Affiliation Reputation}
To assess the reputation of an author's affiliation, we use the average citation count per institution, which serves as a key indicator of an institution’s scholarly influence. Likewise, we map these values into five distinct reputation levels:
\begin{itemize}
    \item Developing Institution (very low): \( 0 \leq \) average citations \( \leq 5 \)
    \item Emerging Institution (low): \( 6 \leq \) average citations \( \leq 15 \)
    \item Established Institution (medium): \( 16 \leq \) average citations \( \leq 30 \)
    \item Reputable Institution (high): \( 31 \leq \) average citations \( \leq 45 \)
    \item World-Class Institution (very high): \( \) average citations \( > 45 \)
\end{itemize}

\paragraph{Journal Reputation}
To assess the reputation of a journal, we use the JCR, which classifies journals into four quartiles based on their impact factor. To enhance interpretability, we map JCR quartiles into human-readable reputation levels:
\begin{itemize}
    \item Top-Level Journal (very high): JCR Q1
    \item High-Level Journal (high): JCR Q2
    \item Moderate-Level Journal (medium): JCR Q3
    \item Low-Level Journal (low): JCR Q4
\end{itemize}

\paragraph{Definition of High Profile}
We define an article as high-profile if its metadata includes at least one leading expert (very high) author, or one world-class (very high) affiliation, or is published in a top-level (very high) journal. This definition aligns with our intuition, as articles produced by renowned researchers, prestigious institutions, or highly reputable venues are generally perceived as more trustworthy. We compute the high profile rate of retraction in each subset by: $| \textit{High Profile Articles}| / | \textit{Retracted Articles}|$. These high-profile cases are hard to detect, which makes the task particularly challenging.

\paragraph{Missing Rate}
In our analysis of the PubMed Retraction dataset, we observed that certain external information may be absent from specific databases. For instance, a journal might not be indexed by the JCR.
To quantify this, we computed the missing rates for each feature within our dataset:
\begin{itemize}
    \item Author Credibility: 10.6\% missing
    \item Affiliation Reputation: 33.4\% missing
    \item Journal Reputation: Journal Indexing: 41.5\% missing
\end{itemize}

% \section{Details of \texttt{Pub-Guard-LLM}}
% \subsection{Prompts}\label{sec:app_prompt}
% Here, we list the prompts used in this work: (1) Explanation Distillation, Table~\ref{tab:prompt_explanation}; 
% (2) Prompt of Detecting Retracted Articles, Table~\ref{tab:query_prompt}; (3) Prompt of Relevance Evaluation, Table~\ref{tab:relevance_prompt};
% (4) Prompt of Relevance Coherence, Table~\ref{tab:coherence_prompt}.

% \subsection{RAG Mode}\label{sec:app_rag}
% We retrieve relevant top-$l$ articles $\mathcal{A}$ (\textit{all legitimate}) to the input article $x$ from PubMed and use them to validate the claims made in $x$.
% In our experiment, the value $l$ is 5 and the PubMed database is MedRAG PubMed\footnote{\url{https://huggingface.co/datasets/MedRAG/pubmed}} with 3.5M articles. The embedding model here is PubMedBERT\footnote{\url{https://huggingface.co/NeuML/pubmedbert-base-embeddings}} and we use cosine similarity to compute the semantic relatedness.

% \subsection{Output Examples}\label{app:expl}
% We provide an input-output example for each application mode:
% Vanilla Reasoning (Table~\ref{tab:example_vanilla}); RAG (Table~\ref{tab:example_rag}); Debate (Table~\ref{tab:example_debate}).

\section{Details of Experiments}
%\subsection{Baselines}
\subsection{Heuristic Features}\label{sec:heuristics}
Since we incorporate external knowledge to represent metadata levels, these fine-grained features can serve as indicators for retraction detection. To leverage this, we design three heuristics based on different metadata attributes. 
The intuition is that articles from prestigious researchers or well-known institutions are generally perceived as more trustworthy.
To classify articles as retracted or not, we establish a threshold-based approach using metadata credibility scores. 
For example, if an article's highest author credibility falls within [\textit{medium, high, very high}], we classify it as legitimate.
Author credibility and affiliation reputation are categorized into six levels (including null). We define the three lower levels as indicators of retraction and the three higher levels as indicators of legitimacy.
Journal reputation, which consists of five levels, is categorized similarly: the three lower levels indicate potential retraction, while the higher levels suggest legitimacy.

\subsection{Evaluation of Distilled Explanation}\label{sec:eval_explanation}
In this evaluation, we provide our co-authors (NLP people and medics) with 
questionnaires. 
This questionnaire is designed to evaluate the quality of explanations generated by an LLM. For each case, the LLM receives information about a medical publication and generates an explanation according to the "Is Retracted" label. This evaluation focuses on this key question ``\textit{ Is the explanation coherent and relevant?
Coherent means the explanation is logical and consistent. Relevant means the explanation is closely connected and appropriate to the provided article information.}''
For each article, our co-authors answer one binary question and can add optional comments if they wish. Finally, we evaluate 45 randomly selected articles and 71.1\% of these explanations are coherent and relevant.

\subsection{User Study}\label{sec:app_user_study}
In the user study, we prepared three questionnaires to evaluate the  %Naive 
Vanilla Reasoning, RAG and Debate modes independently. Using the Vanilla Reasoning mode, we generated explanations for 300 papers. We randomly selected five retracted papers and five legitimate papers from the result and again randomized the order of the paper to create the questionnaire. For each paper, we added 1) the information of the paper as it was in the model prompt; and 2) the explanations generated by the model. Three questions were asked for each paper: 1) whether the explanation was coherent and relevant; 2) %asked the user 
to infer the predicted label from the given explanation; and 3) whether the user agreed with the actual %\FT{actual}
%\todo{or the predicted by the user?} 
prediction. %This process was repeated to create 
% \FT{The same process was followed for}
The same process was followed for the other two questionnaires for the Debate and RAG models, except that, for the Debate model, the information of a supporting viewer and an attacking reviewer was added to the paper description. In total, 30 distinct papers were selected for evaluation by three clinicians.
% \FT{by three clinicians}.

Given the disagreement between human experts and model-generated outputs regarding the debate mode, we present the first three comments below. We hope these critical insights will be valuable for future studies.
\begin{itemize}
    \item \textit{I still don't understand what or who the "attacking reviewer" is. Is this a real world person who has completed a peer review? I feel that labelling this as fraudulent without more detailed evidence of actual fraud is not correct. This is very different from this simply being "not very good" science. I don't think it is fair to effectively discredit an author simply because they have a low H factor. Everyone has to start somewhere. Moreover,   the interpretation does not provide any evidence at all that the authors have either falsified data, that they have a track record of falsifying data, that the authors are affiliated with authors suspected activity or that they have links to governmental or state actors that may want to push falsified data. There is no comment on funding or conflicts of interest which I would expect to be in this analysis and which is missing. The critique argues that "specific" data is missing from the abstract, but then it does not state what this is and what would be required for the abstract to be robust. In summary, this LLM suggests this is research fraud, when as presented, the data suggest that more simply the quality of the abstract is low. This is an important distinction, that could lead to legal proceedings. }
    \item \textit{I think this LLM review is a false positive. I think the criticism of the abstract is not founded. They have performed in vivo and in vitro validation studies, and knock out experiments. While detail is missing, this is an overly harsh critique of a piece of basic science, that is not pretending to be clinical data. Again, what data is missing form the author's institutions? It doesn't actually state which part of this data is important and why. We know they are from a reputable Japanese university, which seems enough to me. Again, to call in to question the credibility of the scientist is libellous. This would need much stronger evidence of criminal or fraudulent behaviour which is clearly missing. So,  I think this is just a case of a not very well written abstract, that has been mis labelled as fraud.}
    \item \textit{The LLM is written in an overly cautious manner, which makes me lack trust in it. Firstly, there is a paradox in its explanation. It states the authors have a high H index, but the institution is emerging... so which are we to judge as being more important? This to me is biasing smaller organisations that may produce outstanding work. We need this model to identify EXCELLENCE in science and to promote it where it exists irrespective of the institution. The factors these models are basing their analysis on e.g. H factor, institutional reputation and author details are not important. I want to understand patterns of publishing behaviours, previous track records of fraud, history of COIs... etc. etc.. and most importantly a robust analysis of the QUALITY of the science. The LLM really provides no insights on this. }
\end{itemize}

\subsection{Ablation Study}
To validate the effect of different components in our methodology, we either vary or remove them and observe their impact on model performance, as measured by F1@Val in Table~\ref{tab:ablation}.
First, we observe that fine-tuning is the most important component, as its removal causes a dramatic performance drop. This indicates it is necessary to adapt LLMs to this task of retraction detection.
External knowledge is very beneficial, and the removal leads to a significant decrease. 
Meanwhile, other components can provide meaningful benefits, which confirms the effectiveness of our methodology.

% \begin{table}[t]
% 	\centering
%     \small
% 	\setlength{\tabcolsep}{1.3mm}{
% 		\begin{threeparttable} 
% 			\begin{tabular}{ccccccc}  
% 				\toprule 
%                      & \textbf{\underline{\texttt{Test}}}  &\textbf{\underline{\texttt{Breast}}} &\textbf{\underline{\texttt{Lung}}}       &\textbf{\underline{\texttt{Ovarian}}}&\textbf{\underline{\texttt{Colorectal}}}&\textbf{\underline{\texttt{Avg}}}\cr
% 				\midrule
%                     Vanilla &84.8&88.5&89.2&85.9&69.7&83.6\cr
%                     RAG & 66.9& 82.7& 87.9 & 85.6&\textbf{81.8}&81.0\cr
%                     Debate & \textbf{90.7} & \textbf{92.2} & \textbf{94.6}& \textbf{89.9}& 75.9&\textbf{88.7} \cr
% 				\bottomrule
% 			\end{tabular}
% 			\caption{%Comparisons of 
%             Recall (average over 3 seeds)  across %three
%             modes } 	\label{tab:recall}
% 		\end{threeparttable}
% 	}
% 	%\end{minipage}%
%  \vspace{-10pt}
% \end{table}

\begin{table}[t] 
	\centering
        \small
	\setlength{\tabcolsep}{1.8mm}{
		\begin{threeparttable} 
			\begin{tabular}{ccc}  
				\toprule 
                    &F1@Val&$\Delta$\cr
                    \midrule
                    Pub-Guard-LLM (\textit{Vanilla})&69.5&-\cr
                    \midrule
                    \multicolumn{3}{c}{\textit{Varying the Base LLM}}\cr
                    \midrule
                    Bio-Llama&68.6&-0.9\cr
                    Mistral&67.8&-1.7\cr
                    \midrule
                    \multicolumn{3}{c}{\textit{Varying the Teacher Model}}\cr
                    \midrule
                    GPT-4o&68.3&-1.2\cr
                    Llama-70B&66.4&-3.1\cr
                    \midrule
                    \multicolumn{3}{c}{\textit{Removing Components}}\cr
                    \midrule
                    \textit{without} Fine-tuning &29.9&-39.6\cr
                    \textit{without} Explanation &68.5&-1.0\cr
                    \textit{without} External Knowledge &59.8&-9.7\cr
                     
				\bottomrule
			\end{tabular}
			\caption{Ablation study results
            } 	\label{tab:ablation}
		\end{threeparttable}
	}
 \vspace{-10pt}
\end{table}

\begin{table*}[ht]
    \centering
    \small
    \renewcommand{\arraystretch}{1.3}
    \setlength{\tabcolsep}{5pt}
    \begin{tabular}{m{14cm}}
        \toprule
        \rowcolor{gray!10} \centering \textbf{Prompt of Explanation Distillation}\cr
        \midrule
        This following article is a \{\texttt{label}\} article.\\
        You are tasked with providing explanations and making a firm case for why it has issues and should be \{\texttt{label}\}.\\ 
        This is the given article:\\
        Title: \{\texttt{Title}\}\\
        Abstract: \{\texttt{Abstract}\}\\
        Authors: \{\texttt{Authors}\}\\
        Affiliations: \{\texttt{Affiliations}\}\\
        Journal: \{\texttt{Journal}\}\\
        Based on the provided information, give a very short and concise explanation why it should be \{\texttt{label}\}\\
        Here are some factors to consider: \\
        (1) the reputation of the journal (whether it is a top journal with a rigorous peer review process)\\
        (2) the reputation of the authors and their affiliations (whether the authors have a history of misconduct in their research)\\
        (3) the integrity of the title and abstract (e.g., controversial topic, using made-up data, plagiarism, etc)\\
        In addition, if check if Email addresses are provided, check if it conforms to institutional format. Otherwise, no need to make comments about the Email adresses.\\
        Only provide the key reason briefly (in \{\texttt{Token}\} words) and do not repeat the title and sentences of abstract.\\
        Note that we use the token 'null' to indicate that additional information is missing from the corresponding databases.\\
        Write a separate sentence for each feature to avoid creating long and complex sentences.\\
        \bottomrule
    \end{tabular}
    \caption{Example prompt used for distilling explanations}
    \label{tab:prompt_explanation}
\end{table*}

\begin{table*}[ht]
    \centering
    \small
    \renewcommand{\arraystretch}{1.3}
    \setlength{\tabcolsep}{5pt}
    \begin{tabular}{m{14cm}}
        \toprule
        \rowcolor{gray!10} \centering \textbf{Prompt of Detecting Retracted Articles}\cr
        \midrule
        You are tasked with determining whether a given research paper should be retracted.\\
        To make this judgment, analyze the provided title, abstract, author information, 
        institutional affiliation, and publishing journal.\\
        Here are some factors to consider: \\
        (1) the reputation of the journal (whether it is a top journal with a rigorous peer review process)\\
        (2) the reputation of the authors and their affiliations (whether the authors have a history of misconduct in their research)\\
        (3) the integrity of the title and abstract (e.g., controversial topic, using made-up data, plagiarism, etc)\\
        In addition, if check if Email addresses are provided, check if it conforms to institutional format. Otherwise, no need to make comments about the Email adresses.\\
        
        ---------------------------------------\\
        Use the examples below as guidance:\\
        Example:\\
        Title: \{\texttt{Title}\}\\
        Abstract: \{\texttt{Abstract}\}\\
        Authors: \{\texttt{Authors}\}\\
        Affiliations: \{\texttt{Affiliations}\}\\
        Journal: \{\texttt{Journal}\}\\
        ---------------------------------------\\
        
        This is the given article:\\
        Title: \{\texttt{Title}\}\\
        Abstract: \{\texttt{Abstract}\}\\
        Authors: \{\texttt{Authors}\}\\
        Affiliations: \{\texttt{Affiliations}\}\\
        Journal: \{\texttt{Journal}\}\\
        Please first provide your prediction of whether this paper should be retracted and then provide your assessment and explanation.\\
        Label (answer Yes or No): \\
   
        \bottomrule
    \end{tabular}
    \caption{The prompt used to query LLMs.}
    \label{tab:query_prompt}
\end{table*}

\begin{table*}[ht]
    \centering
    \small
    \renewcommand{\arraystretch}{1.3}
    \setlength{\tabcolsep}{5pt}
    \begin{tabular}{m{14cm}}
        \toprule
        \rowcolor{gray!10} \centering \textbf{Prompt of Relevance Evaluation}\cr
        \midrule
        Evaluate the relevance of explanation provided for the model's prediction of the retracted articles.\\ 
        Consider the following criteria:\\
        Fidelity: Does the explanation highlight specific features and patterns within the article?\\
        Accuracy: Does the explanation avoid mentioning elements that do not exist in the article? The model should not make hallucinated explanations.\\
        Based on these criteria, rate the explanation's relevance on a scale from 1 to 10.\\
        -----------------------------------\\
        This is the given article:\\
        Title: \{\texttt{Title}\}\\
        Abstract: \{\texttt{Abstract}\}\\
        Authors: \{\texttt{Authors}\}\\
        Affiliations: \{\texttt{Affiliations}\}\\
        Journal: \{\texttt{Journal}\}\\
         -----------------------------------\\
        This is the prediction and explanation:\\
        Prediction and Explanation: \{\texttt{explanation}\}\\
        Rate the relevance, only provide the score:\\
         
        \bottomrule
    \end{tabular}
    \caption{Example prompt used for evaluating the relevance of explanations}
    \label{tab:relevance_prompt}
\end{table*}

\begin{table*}[ht]
    \centering
    \small
    \renewcommand{\arraystretch}{1.3}
    \setlength{\tabcolsep}{5pt}
    \begin{tabular}{m{14cm}}
        \toprule
        \rowcolor{gray!10} \centering \textbf{Prompt of Coherence Evaluation}\cr
        \midrule
        Evaluate the coherence of the explanation provided for the model's prediction of retracted articles.
        Consider the following criteria:\\
        Coherence: Does the explanation maintain logical consistency and clarity in its reasoning?\\
        Structure: Is the explanation well-organized, with ideas flowing logically from one to another?\\
        Based on these criteria, rate the explanation's coherence on a scale from 1 to 10.\\
        -----------------------------------\\
        This is the given article:\\
        Title: \{\texttt{Title}\}\\
        Abstract: \{\texttt{Abstract}\}\\
        Authors: \{\texttt{Authors}\}\\
        Affiliations: \{\texttt{Affiliations}\}\\
        Journal: \{\texttt{Journal}\}\\
         -----------------------------------\\
        This is the prediction and explanation:\\
        Prediction and Explanation: \{\texttt{explanation}\}\\
        Rate the coherence, only provide the score:\\
         
        \bottomrule
    \end{tabular}
    \caption{Example prompt used for evaluating the coherence of explanations}
    \label{tab:coherence_prompt}
\end{table*}

\begin{table*}[ht]
    \centering
    \small
    \renewcommand{\arraystretch}{1.3}
    \setlength{\tabcolsep}{5pt}
    \begin{tabular}{m{14cm}}
        \toprule
        \rowcolor{gray!10} \centering \textbf{An Example of the Vanilla Reasoning Mode}\cr
        \midrule
        \textbf{User Query with External Knowledge:}\\
        \textit{Title:} Changes and Influencing Factors of Cognitive Impairment in Patients with Breast Cancer\\
        \textit{Abstract:} To investigate the changes in cognitive function and its influencing factors in patients with breast cancer after chemotherapy,
        to provide a scientific basis for further cognitive correction therapy. In this study, general information on age,
        marital status, and chemotherapy regimen was collected from 172 breast cancer chemotherapy patients.
        172 patients with breast cancer undergoing chemotherapy were investigated by convenience sampling method,
        and the subjects were tested one-on-one using the Chinese version of the MATRICS Consensus Cognitive Battery (MCCB) computer system.\\

        \textit{Authors:} Cui (author h-index: 6, Early Career Researcher);
        Song (author h-index: 1, Emerging Researcher); Zhang (author h-index: 7, Early Career Researcher)\\

        \textit{Institutions:} College of Nursing, Jinzhou Medical University, Jinzhou, Liaoning 121001, China. (institution average citation: 9.0, Emerging Institution); Department of Intensive Care Medicine, Liaocheng People's Hospital, Liaocheng, Shandong 252000, China. (institution average citation: 10.0, Emerging Institution);\\

        \textit{Journal:} evidence-based complementary and alternative medicine : ecam (null)\\

        \textbf{Pub-Guard-LLM:}\\
        \textit{Prediction:} \\
        Yes (\textit{Retracted})\\ 
        \textit{Explanation:}\\
         The article should be retracted due to concerns about the integrity of the study. The authors' affiliations are inconsistent, with some listed multiple times, suggesting a potential error or lack of proper collaboration. The journal in which it was published is not specified, making it difficult to assess the rigor of the peer review process. The study's sample size is relatively small, which may limit the generalizability of the findings. Additionally, the abstract does not provide sufficient detail about the methods used, making it difficult to assess the validity of the results.\\
        
        \bottomrule
    \end{tabular}
    \caption{An input-output example of the Vanilla Reasoning mode}
    \label{tab:example_vanilla}
\end{table*}

\begin{table*}[ht]
    \centering
    \small
    \renewcommand{\arraystretch}{1.3}
    \setlength{\tabcolsep}{5pt}
    \begin{tabular}{m{14cm}}
        \toprule
        \rowcolor{gray!10} \centering \textbf{An Example of the RAG Mode}\cr
        \midrule
        \textbf{User Query with External Knoledge:}\\
        \textit{Title:} Changes and Influencing Factors of Cognitive Impairment in Patients with Breast Cancer\\
        \textit{Abstract:} To investigate the changes in cognitive function and its influencing factors in patients with breast cancer after chemotherapy,
        to provide a scientific basis for further cognitive correction therapy. In this study, general information on age,
        marital status, and chemotherapy regimen was collected from 172 breast cancer chemotherapy patients.
        172 patients with breast cancer undergoing chemotherapy were investigated by convenience sampling method,
        and the subjects were tested one-on-one using the Chinese version of the MATRICS Consensus Cognitive Battery (MCCB) computer system.\\

        \textit{Authors:} Cui (author h-index: 6, Early Career Researcher);
        Song (author h-index: 1, Emerging Researcher); Zhang (author h-index: 7, Early Career Researcher)\\

        \textit{Institutions:} College of Nursing, Jinzhou Medical University, Jinzhou, Liaoning 121001, China. (institution average citation: 9.0, Emerging Institution); Department of Intensive Care Medicine, Liaocheng People's Hospital, Liaocheng, Shandong 252000, China. (institution average citation: 10.0, Emerging Institution);\\

        \textit{Journal:} evidence-based complementary and alternative medicine : ecam (null)\\

        \textbf{RAG Results:}\\
        In the meantime, a set of legitimate papers is provided, each containing highly relevant articles.
        Use these articles as evidence to assess whether the given paper is potentially retracted.
        Look for indicators such as controversial topics, unsupported claims, or inconsistencies.
        Below are examples of relevant papers:\\
        Example 1:\\
        Title: \{\texttt{Title}\}
        Abstract: \{\texttt{Abstract}\}\\
        ......\\
        
        \textbf{Pub-Guard-LLM:}\\
        \textit{Prediction:} \\
        Yes (\textit{Retracted})\\ 
        \textit{Explanation:}\\
         The article should be retracted due to the lack of reputation of the journal, as it is not listed in the provided information. Additionally, the authors have relatively low h-index scores, indicating a lack of established reputation in their field. Furthermore, the study's sample size is small, and the results may not be generalizable to a larger population. Lastly, the article lacks information on the study's methodology and statistical analysis, raising concerns about the validity of the findings\\
        
        \bottomrule
    \end{tabular}
    \caption{An input-output example for the RAG mode}
    \label{tab:example_rag}
\end{table*}

\begin{table*}[ht]
    \centering
    \small
    \renewcommand{\arraystretch}{1.3}
    \setlength{\tabcolsep}{5pt}
    \begin{tabular}{m{14cm}}
        \toprule
        \rowcolor{gray!10} \centering \textbf{An Example of the Debate Mode}\cr
        \midrule
        \textbf{User Query with External Knowledge:}\\
        \textit{Title:} Changes and Influencing Factors of Cognitive Impairment in Patients with Breast Cancer\\
        \textit{Abstract:} To investigate the changes in cognitive function and its influencing factors in patients with breast cancer after chemotherapy,
        to provide a scientific basis for further cognitive correction therapy. In this study, general information on age,
        marital status, and chemotherapy regimen was collected from 172 breast cancer chemotherapy patients.
        172 patients with breast cancer undergoing chemotherapy were investigated by convenience sampling method,
        and the subjects were tested one-on-one using the Chinese version of the MATRICS Consensus Cognitive Battery (MCCB) computer system.\\

        \textit{Authors:} Cui (author h-index: 6, Early Career Researcher);
        Song (author h-index: 1, Emerging Researcher); Zhang (author h-index: 7, Early Career Researcher)\\

        \textit{Institutions:} College of Nursing, Jinzhou Medical University, Jinzhou, Liaoning 121001, China. (institution average citation: 9.0, Emerging Institution); Department of Intensive Care Medicine, Liaocheng People's Hospital, Liaocheng, Shandong 252000, China. (institution average citation: 10.0, Emerging Institution);\\

        \textit{Journal:} evidence-based complementary and alternative medicine : ecam (null)\\

        \textbf{Supporting Reviewer:}\\
        The article appears legitimate based on the following factors:\\
        1. Journal Reputation: The article is published in evidence-based complementary and alternative medicine : ecam, a journal dedicated to complementary and alternative medicine, suggesting a focus on rigorous scientific research in this field.
        2. Author Reputation: The authors are affiliated with recognized institutions, and their h-index scores, though relatively low, indicate they have published and been cited in other academic works.
        3. Title and Abstract Integrity: The title and abstract present a clear research question, methodology, and findings, with no apparent signs of controversial topics, made-up data, or plagiarism.
        4. Institutional Affiliation: The authors are associated with various departments and colleges within recognized universities, suggesting a multidisciplinary approach to the research.
        5. Missing Information: The 'null' values indicate missing information, but this does not necessarily delegitimize the article, as it may simply be a database issue.\\

        \textbf{Attacking Reviewer:}\\
        The article should be retracted due to the lack of reputation of the journal, as it is not specified and could indicate a lack of rigorous peer review. The authors have relatively low h-indices, suggesting they are early in their careers or have not yet produced highly impactful research. The study's findings may not be reliable as the sample size is not specified, and the results could be biased due to the convenience sampling method used. Additionally, the absence of information about the Department of Radiation Oncology at Jinzhou Medical University raises concerns about the credibility of the research.\\
        
        \textbf{Pub-Guard-LLM:}\\
        \textit{Prediction:} \\
        Yes (\textit{Retracted})\\ 
        \textit{Explanation:}\\
         The article should be retracted due to concerns about the credibility of the research. The journal's reputation is questionable as it is not specified, suggesting a lack of rigorous peer review. The authors' relatively low h-indices and the absence of information about one of the institutions raise questions about their expertise and the validity of the research. Furthermore, the study's findings may not be reliable due to the lack of information about the sample size and the use of convenience sampling, which could introduce bias.\\
        
        \bottomrule
    \end{tabular}
    \caption{An input-output example for the Debate mode}
    \label{tab:example_debate}
\end{table*}

\end{document}